# Discriminative Probabilistic Prototype Learning

**Edwin V. Bonilla**                                    edwin.bonilla@nicta.com.au

NICTA & Australian National University, Locked Bag 8001, Canberra ACT 2601, Australia

**Antonio Robles-Kelly**                          Antonio.Robles-Kelly@nicta.com.au

NICTA & Australian National University, Locked Bag 8001, Canberra ACT 2601, Australia

## Abstract

In this paper we propose a simple yet powerful method for learning representations in supervised learning scenarios where an input datapoint is described by a set of feature vectors and its associated output may be given by soft labels indicating, for example, class probabilities. We represent an input datapoint as a K-dimensional vector, where each component is a mixture of probabilities over its corresponding set of feature vectors. Each probability indicates how likely a feature vector is to belong to one-out-of-K unknown prototype patterns. We propose a probabilistic model that parameterizes these prototype patterns in terms of hidden variables and therefore it can be trained with conventional approaches based on likelihood maximization. More importantly, both the model parameters and the prototype patterns can be learned from data in a *discriminative* way. We show that our model can be seen as a probabilistic generalization of learning vector quantization (LVQ). We apply our method to the problems of shape classification, hyperspectral imaging classification and people's work class categorization, showing the superior performance of our method compared to the standard prototype-based classification approach and other competitive benchmarks.

## 1. Introduction

A fundamental problem in machine learning is that of coming up with useful characterizations of the input so that we can achieve better generalization capabilities with our learning algorithms. We refer to this problem as that of *learning representations*. This has been a long standing goal in machine learning and has been addressed throughout the years from different perspectives. In fact, one of the simplest and oldest attempts to tackle this problem was given by Rosenblatt (1962) with the Perceptron algorithm for classification problems. In such an approach it was suggested that we can have non-linear mappings of the inputs so that the obtained representation allows us to discriminate between the classes with a simple linear function. However, the mapping (or "features") should have been engineered beforehand instead of being learned from the available data. Neural networks and their back-propagation algorithm (Rumelhart et al., 1986) became popular because they offered an automatic way of learning flexible representations by introducing the so-called hidden layers and hidden units into multilayer Perceptron architectures. Kernel-based algorithms and, in particular, support vector machines (SVMs, Scholkopf & Smola, 2001) offered a clever alternative to neural nets, circumventing the problem of learning representations by using kernels to map the input into feature spaces where the patterns are likely to be linearly separable. However, SVMs are inherently non-probabilistic and unsuitable for applications where one requires uncertainty measures around their predictions.

In this paper we propose a simple yet powerful approach to learning representations for classification problems where an original input datapoint is described by a set of feature vectors and its associated output may be given by soft labels indicating, for example, class probabilities, degrees of membership or noisy labels. Our approach to this problem is to represent an input datapoint as a K-dimensional vector, where each component is a mixture of probabilities over its corresponding set of feature vectors. Each probability indicates how likely a feature vector is to belong to one-out-of-K unknown prototype patterns.

We propose a probabilistic model that parameterizes these prototype patterns in terms of hidden variables and therefore it can be trained with conventional approaches based on likelihood maximization. More importantly, both the model parameters and the prototype patterns can be learned from data in a *discriminative* way. To our knowledge, previous approaches have not addressed the problem of discriminative prototype learning within a consistent multi-class probabilistic framework (see section 5 for details).

## 2. Problem Setting

In this paper we are interested in multi-class classification problems for which an input point is characterized by a set of feature vectors $\mathcal{S} = \{\mathbf{x}^{(1)}, \dots, \mathbf{x}^{(M)}\}$, where each feature vector may describe, for example, some local characteristics of the input point. Additionally, we consider the general case where the outputs may be given by soft labels indicating, for example, class probabilities, degrees of membership or noisy class labels. Hence, our goal is to build a probabilistic classifier based on given training data, which is comprised by the tuples $\mathcal{D} = \{(\mathcal{S}^{(n)}, \tilde{\mathcal{P}}^{(n)}), n = 1, \dots, N\}$, where $\mathcal{S}^{(n)}$ is the set of feature vectors in the $n^{\text{th}}$ tuple. In general, the number of feature vectors is different for each input and therefore we shall describe it by $M_n$. Similarly, $\tilde{\mathcal{P}}^{(n)} \in [0,1]^C$, corresponds to the C-dimensional vector of soft labels (e.g. empirical probabilities) associated with the C output classes of the $n^{\text{th}}$ training instance. Obviously, these probabilities are constrained by $\sum_{j=1}^{C} \tilde{P}(y^n = j) = 1$, where $y^n$ denotes the latent class assignment of datapoint $n$.

A common approach to prototype-based learning describes an input by a histogram of words from a vocabulary of size $K$. This histogram is commonly known as the bag-of-words representation. In order to specify the vocabulary it is customary to use clustering methods such as $K$-means or generative models such as Gaussian Mixtures, which are often used as a disjoint step before training a specific classifier. The model for extracting such representations is given by:

$$f_k(\mathbf{x}) = \begin{cases} 1 \text{ iff } \|\boldsymbol{\mu}_k - \mathbf{x}\| < \|\boldsymbol{\mu}_j - \mathbf{x}\| \ \forall j \neq k \\ 0 \text{ otherwise} \end{cases} \quad (1)$$

$$z_k^{(n)} = \sum_{\mathbf{x} \in \mathcal{S}^{(n)}} \pi_k f_k(\mathbf{x}), \quad (2)$$

where $\mathbf{z}$ is a $K$-dimensional vector to be used as the input representation for a specific classifier; $\{\boldsymbol{\mu}_k\}_{k=1}^K$ are usually referred to as the centers; and $\pi_k$ is set to $1/M_n$. We will refer to each $f_k(\mathbf{x})$ as a prototype function as it performs the encoding of each $D$-dimensional vector $\mathbf{x}$ into its corresponding (binary) $K$-dimensional representation. It is important to realize that this method is a winner-takes-all approach where each feature vector $\mathbf{x}$ is assigned to only one centre. This is the main motivation for our method where we will relax this assumption and propose a fully discriminative probabilistic model for learning such representations.

Our model assumes a bag-of-words representation as given by equation (2). However, here we consider that the prototypes are *probabilities* and are given by:

$$f_k(\mathbf{x}) = \frac{\exp(-\beta\|\boldsymbol{\mu}_k - \mathbf{x}\|^2)}{\sum_{j=1}^{K} \exp(-\beta\|\boldsymbol{\mu}_j - \mathbf{x}\|^2)}, \quad (3)$$

where $\beta$ is a rate parameter (or inverse temperature). Note that when $\beta \to \infty$ Equation (3) becomes equivalent to the hard limit in Equation (1). Therefore, $f_k(\mathbf{x})$ is the probability of feature vector $\mathbf{x}$ belonging to "cluster" $k$ and $z_k$ is a mixture of these probabilities.

In addition to defining how to map the set of input vectors into parameterized probabilistic prototype representations, our method requires the definition of a discriminative probabilistic classifier. In principle, this could be any classifier that focuses on defining the conditional probability:

$$\sigma^i(\mathbf{z}(\mathbf{X}); \boldsymbol{\Theta}) \overset{\text{def}}{=} p(y = i|\mathbf{z}(\mathbf{X}), \boldsymbol{\Theta}) \quad (4)$$

directly in terms of our prototype representation $\mathbf{z}$. Here we have used $\mathbf{X} \overset{\text{def}}{=} \{\mathbf{x}^{(j)}\}_{j=1}^M$ and made explicit the dependency of $\mathbf{z}$ on its corresponding feature vectors. In the sequel, for simplicity in the notation, we will drop this dependency. Note that our model is a (conditional) directed probabilistic model. This contrasts with other approaches such as latent-variable CRFs which are undirected graphical models. As we shall see later, we will focus on a softmax classifier due to the simplicity and efficacy of this parametric model. However, it is clear that we can also incorporate non-parametric classifiers. We expect such approaches to be more effective than their parametric counterparts and we postpone their study to future work.

## 3. Parameter Learning

In this section we are interested in learning the parameters of our discriminative probabilistic prototype framework. These parameters are: the rate parameter $\beta$, the vocabulary or centers $\{\boldsymbol{\mu}_k\}_{k=1}^K$ and the parameters of the discriminative classifier $\boldsymbol{\Theta}$. This could be effected using a number of optimization methods including simulated annealing and Markov Chain Monte Carlo. Here we propose direct gradient-based optimization of the data log-likelihood.

### 3.1. Direct Likelihood Maximization with Gradient-Based Methods

Assuming iid data, the log-likelihood of the model parameters given the data can be expressed as:

$$\mathcal{L}(\boldsymbol{\Theta}, \{\boldsymbol{\mu}_k\}_{k=1}^k, \beta) = \sum_{n=1}^{N} \mathcal{L}^n(\boldsymbol{\Theta}, \{\boldsymbol{\mu}_k\}_{k=1}^K, \beta) \quad (5)$$

$$= \sum_{n=1}^{N} \tilde{P}(y^n) \log P\left(y^n | \mathbf{z}^n(\boldsymbol{\Theta}, \{\boldsymbol{\mu}_k\}_{k=1}^K, \beta)\right), \quad (6)$$

where $\tilde{P}(y^n)$ refers to the soft labels associated with input $n$, e.g. the empirical class probabilities. In order to optimize the data log-likelihood we use a Quasi-Newton method (BFGS) to which we provide the following gradient information:

$$\nabla_{\boldsymbol{\mu}^\ell} \mathcal{L} = \sum_{n=1}^{N} \nabla_{\boldsymbol{\mu}^\ell} \mathcal{L}^n \qquad \frac{\partial \mathcal{L}}{\partial \beta} = \sum_{n=1}^{N} \frac{\partial \mathcal{L}^n}{\partial \beta} \quad (7)$$

and

$$\nabla_{\boldsymbol{\mu}^\ell} \mathcal{L}^n = (\mathbf{G}^{(n,\ell)})^T \mathbf{g}^{(n)}, \quad (8)$$

where:

$$\mathbf{g}^{(n)} \stackrel{\text{def}}{=} \nabla_{\mathbf{z}^n} \mathcal{L}^n \qquad \mathbf{G}^{(n,\ell)} \stackrel{\text{def}}{=} \nabla_{\boldsymbol{\mu}^\ell} \mathbf{z}^n \quad (9)$$

$$g_k^{(n)} = \frac{\partial \mathcal{L}^n}{\partial z_k^n} \qquad G_{k,d}^{(n,\ell)} = \frac{\partial z_k^n}{\partial \mu_d^\ell} \quad (10)$$

for $k, \ell = 1, \dots K$ and $d = 1, \dots D$. The advantage of the above formulation is that the partial derivatives wrt the cluster centers can be computed in closed-form. Note that although these updates are straightforward to derive, we specify all the details here for completeness and for their later use when showing the relation between our method and LVQ in Section 3.3. The derivatives are given as follows:

$$\frac{\partial \mathcal{L}^n}{\partial \mu_d^\ell} = \sum_{k=1}^{K} \frac{\partial \mathcal{L}^n}{\partial z_k^n} \frac{\partial z_k^n}{\partial \mu_d^\ell}. \quad (11)$$

Moreover, by using equations (2) and (3) we have that:

$$\nabla_{\boldsymbol{\mu}^\ell} z_k^n = \sum_{\mathbf{x} \in \mathcal{S}^n} \nabla_{\boldsymbol{\mu}^\ell} f_k(\mathbf{x}) \quad (12)$$

$$= \begin{cases} 2\beta \sum_{\mathbf{x} \in \mathcal{S}^n} f_k(\mathbf{x})(1 - f_k(\mathbf{x}))(\mathbf{x} - \boldsymbol{\mu}^\ell), \text{ iff } k = \ell \\ 2\beta \sum_{\mathbf{x} \in \mathcal{S}^n} f_k(\mathbf{x}) f_\ell(\mathbf{x})(\boldsymbol{\mu}^\ell - \mathbf{x}), \text{ otherwise.} \end{cases} \quad (13)$$

A similar approach can be followed to compute the partial derivatives wrt $\beta$. Hence we have that:

$$\frac{\partial \mathcal{L}^n}{\partial \beta} = \sum_{k=1}^{K} \frac{\partial \mathcal{L}^n}{\partial z_k^n} \frac{\partial z_k^n}{\partial \beta} \quad (14)$$

$$\frac{\partial z_k^n}{\partial \beta} = \sum_{\mathbf{x} \in \mathcal{S}^n} f_k(\mathbf{x}) \left( \sum_{\ell=1}^{K} \|\boldsymbol{\mu}^\ell - \mathbf{x}\|^2 f_\ell(\mathbf{x}) - \|\boldsymbol{\mu}^k - \mathbf{x}\|^2 \right). \quad (15)$$

### 3.2. Discriminative Parametric Model: Softmax Classifier

For the case of a parametric model such as the softmax classifier:

$$P(y = i | \mathbf{z}, \boldsymbol{\Theta}) = \frac{\exp((\boldsymbol{\theta}^i)^T \mathbf{z})}{\sum_{j=1}^{C} \exp((\boldsymbol{\theta}^j)^T \mathbf{z})} \stackrel{\text{def}}{=} \sigma^i(\mathbf{z}; \boldsymbol{\Theta}) \quad (16)$$

the local log-likelihood terms can be written as:

$$\mathcal{L}^n = \sum_{j=1}^{C} \tilde{P}(y^n = j) \log \sigma^j(\mathbf{z}^n; \boldsymbol{\Theta}). \quad (17)$$

The corresponding derivatives wrt the prototype representations are given by:

$$\nabla_{\mathbf{z}^n} \mathcal{L}^n = \sum_{j=1}^{C} \left( \tilde{P}(y^n = j) - \sigma^j(\mathbf{z}^n; \boldsymbol{\Theta}) \right) \boldsymbol{\theta}^j. \quad (18)$$

Finally, we also need the gradient information wrt the model parameters ($\boldsymbol{\Theta}$):

$$\nabla_{\boldsymbol{\theta}_i} \mathcal{L} = \sum_{n=1}^{n} \left( \tilde{P}(y^n = i) - \sigma^i(\mathbf{z}^n; \boldsymbol{\Theta}) \right) \mathbf{z}^n. \quad (19)$$

Equations (5) and (19) can be modified so as to include a regularization term $-\lambda \operatorname{tr}(\boldsymbol{\Theta}^T \boldsymbol{\Theta})$, where $\lambda$ is a regularization parameter; $\operatorname{tr}(\cdot)$ is the trace operator; and $\boldsymbol{\Theta}$ is the matrix of weights with columns given by each $\boldsymbol{\theta}^i$, $i = 1 \dots, C$. It is well known that the solution for the weights in this case corresponds to the MAP solution when considering a Gaussian prior over the weights $\boldsymbol{\theta}_i \sim \mathcal{N}(\boldsymbol{\theta}_i | \mathbf{0}, \lambda \mathbf{I})$. We can have a similar prior or regularization for $\beta$.

We note here that optimization of the objective function in Equation (5) wrt to all the parameters is a non-convex problem. However, as we shall see in section 4, we use coordinate ascent and optimization of the model parameters $\boldsymbol{\Theta}$, given all others fixed, is a convex problem.

### 3.3. Relation to LVQ

Learning Vector Quantization (LVQ, Kohonen, 1990) is a prototype-based learning algorithm that uses the class label information to adapt the prototypes (i.e. cluster centers). The idea is that the prototypes should move towards the training examples in their corresponding class and away from training examples with different labels. As in the k-means algorithm, the assignment of datapoints to prototypes corresponds to the rule in Equation (1). If we were going to update the prototypes in our model with gradient ascent, this update would become:

$$\boldsymbol{\mu}_\ell^{(\text{new})} = \boldsymbol{\mu}_\ell^{(\text{old})} + \eta \nabla_{\boldsymbol{\mu}_\ell} \mathcal{L}, \qquad (20)$$

where $\eta$ is the learning rate.

By expanding Equation (11), substituting Equations (12) and (18), and assuming hard labels, i.e. $\tilde{P}(y^n)$ is 1 for only one label and zero for all the others, we should get:

$$\boldsymbol{\mu}^{\ell(\text{new})} = \boldsymbol{\mu}^\ell + $$
$$\eta \left[ \sum_{\mathbf{x} \in \mathcal{S}^n} \sum_{k=1}^{K} \left( \theta_k^{y^n} - \sum_{j=1}^{C} \theta_k^j \sigma^j(\mathbf{z}; \boldsymbol{\theta}) \right) f_k(\mathbf{x}) f_\ell(\mathbf{x}) (\boldsymbol{\mu}^\ell - \mathbf{x}) \right.$$
$$\left. + \sum_{\mathbf{x} \in \mathcal{S}^n} \left( \theta_\ell^{y^n} - \sum_{j=1}^{C} \theta_\ell^j \sigma^j(\mathbf{z}; \boldsymbol{\theta}) \right) f_\ell(\mathbf{x})(\boldsymbol{\mu}^\ell - \mathbf{x}) \right]. \qquad (21)$$

By taking the hard limit for the assignment of the samples to a single prototype (and assuming that all samples corresponding to a single data-point are assigned to the same cluster): $f_\ell(\mathbf{x}) = 1$ and, consequently, $f_k(\mathbf{x}) = 0$ for $k \neq \ell$ we have:

$$\boldsymbol{\mu}^{\ell(\text{new})} = \boldsymbol{\mu}^\ell + \eta \sum_{\mathbf{x} \in \mathcal{S}^n} \left( \theta_\ell^{y^n} - \sum_{j=1}^{C} \theta_\ell^j \sigma^j(\mathbf{z}; \boldsymbol{\theta}) \right) (\boldsymbol{\mu}^\ell - \mathbf{x}). \qquad (22)$$

If $|S^n| = 1$ then the factor premultiplying $(\boldsymbol{\mu}^\ell - \mathbf{x})$ can be absorbed into the learning rate and, therefore, we obtain the LVQ update. In our model, this factor is interpreted as the difference between the parameter corresponding to the prototype $\ell$ for the label of the current datapoint , i.e $(\theta_\ell^{y^n})$ and the average (wrt the posterior probabilities) of the parameters of the other classes. As a result, we can view the gradient-ascent updates for our method as a relaxed version of LVQ.

## 4. Experiments and Results

In this section we present results on synthetic data, shape classification, hyperspectral image classification and people's work class categorization. For all our experiments, we employ our prototype learning approach where we iterate the learning of the prototype parameters $\{\boldsymbol{\mu}_\ell\}_{\ell=1}^{K}$, $\beta$ and the model parameters $\boldsymbol{\Theta}$ via coordinate ascent on the model likelihood. We have initialized the cluster centers making use of k-means and varied the size of the vocabulary, i.e. $K$, according to the experimental vehicle. To illustrate the behavior of our algorithm and to show its performance with respect to competitive benchmarks, the first three problems studied (synthetic data, shape classification, hyperspectral image classification) consider the common case of hard labels and our last experiment on people's work class categorization investigates the use of class probabilities as soft labels.

### 4.1. Illustration on Synthetic Data

These experiments are based on a set of 20 datapoints each comprised by a set of $M_n \in \{1, \ldots, 20\}$ feature vectors in a 2-D space. Here we compare the baseline model yielded by k-means clustering (with $K = 2$) against our discriminative model. Figure 1 shows the original datapoints in the two-dimensional space along with the cluster centers (top). We see that the cluster centers learned by our discriminative approach (Panel b) are quite different even though the clusters obtained by k-means (Panel a) are very close to those used to generate the data. At the bottom panel we show the prototype representation given by both methods and we observe that the representation learned by our method is much more discriminative as it takes into consideration the class labels.

### 4.2. MPEG DataSet

Our first real dataset is given by the MPEG-7 CE-1 Part B database (Latecki et al., 2000), which we will refer to as the MPEG-7 dataset. This contains 1400 binary shapes organized in 70 classes, each comprised of 20 images. We have sampled 1 in every 10 pixels on the shape contours and we have built a fully connected graph whose edge-weights are given by the Euclidean distances between each pair of pixel locations. These weights are then normalized to be in the interval $[0, 1]$. The feature vectors are given by the frequency histograms of these distances for every node. In our experiments, we have used 10 bins for the frequency histogram computation and set $K = 200$.

For purposes of shape categorization, we compare our method to three alternatives. The first one is a prototype-based baseline akin to the bag of words approaches in computer vision (Fei-Fei & Perona, 2005). Our baseline recovers prototypes via k-means cluster-ing. The shapes under study are then described by the prototype-based representation which we then employ as input to a classifier. The classifier used for this baseline is also a softmax classifier as in our method. In other words, the only difference between this baseline method and our approach (probabilistic prototype) is how these prototypes are learned.

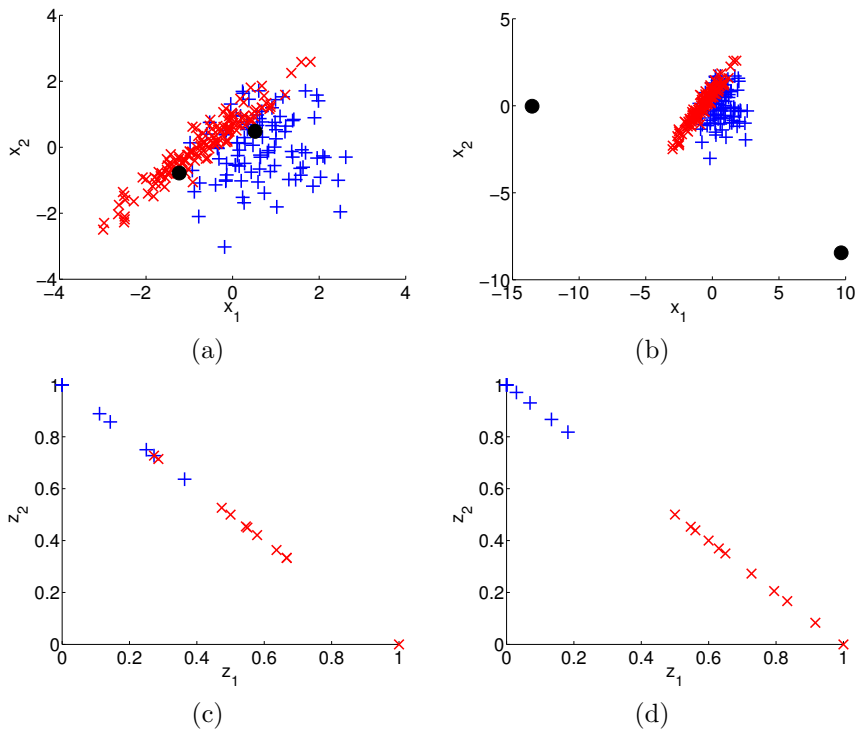

*Figure 1.* Illustration of our model's discriminative power on toy data. (a) The original two-dimensional data drawn from a mixture of Gaussians with two components: one is an isotropic Gaussian with variance 1 and the other one is a correlated Gaussian with covariance 0.95 and variance 1 on both dimensions. An input is represented by several datapoints in the plot and the cluster centers as learned by k-means are shown as black dots. (b) The same data with the cluster centers as learned by our discriminative probabilistic prototype classifier. (c) The representation obtained when using a standard prototype-based approach (k-means). (d) The representation learned when using our discriminative probabilistic prototype model.

The other two alternatives are specifically designed for purposes of shape classification. These are the shape and skeletal contexts in Belongie et al. (2002) and Xie et al. (2008), respectively. Once the shape and skeletal contexts are at hand, we train one-versus-all SVM classifiers whose parameters have been selected through ten-fold cross validation. For all our shape categorization experiments, we have divided the graphs in the dataset into a training and a testing set. The training set comprises 50% randomly selected graphs, i.e. 700 and the other 50% of the data was used for testing. For these partitions we have effected five trials. The categorization results are shown in Table 1. The table shows the mean percentage of correctly classified shapes and the corresponding variance. Despite the basic strategy taken for the construction of our graphs, which contrasts with the specialized nature of the skeletal and shape contexts, our probabilistic prototype method outperforms the alternatives.

## 4.3. SPECTRAL Dataset

This application considers hyperspectral imagery from real-world scenes that include various types of materials. We have annotated these images at a pixel-level considering 10 different classes (C=10): tree trunk, light poles, shadow on grass, grass, road, white line on road, shadow on road, leaves, sky, and white regions on sky. We considered 24 different images from which we have extracted 1,746,708 data points with their corresponding labels. In order to characterize an input data point with a set of vectors ($\mathcal{S}^n$) we have considered neighborhood information according to $7 \times 7$ windows. Hence, for each data point we have 49 feature vectors. We have subsampled the data so as to include 1000 training instances and 16643 test instances across 10 different partitions.

*Table 1.* Shape categorization results for the MPEG-7 dataset. The mean classification accuracy is shown along with (±) one standard deviation when using 50% of the data for training and the rest for testing. Our probabilistic prototype approach outperforms all other baseline methods.

| Probabilistic Prototype | Standard Prototype | Shape Context (Belongie et al., 2002) | Skeletal Context (Xie et al., 2008) |
|---|---|---|---|
| **85.49 ± 1.43 %** | 82.20 ± 0.98 % | 77.55 ± 2.39 % | 79.91 ± 1.78 % |

Note that our method is effectively learning the prototypes that can be used to represent the spectra for the objects in the scene as a linear combination of the spectral signatures of its material constituents. In geosciences and process control this is known as unmixing (Bergman, 2006). Current unmixing methods assume availability of the end-member spectra, which usually involves a cumbersome labeling of the end member data, effected through expert intervention.

For comparison we use the standard prototype approach (based on k-means for learning the centers), and the *InfoLoss* (Lazebnik & Raginsky, 2009) method. InfoLoss adapts the prototypes based on the optimization of an information-theoretical criterion. For this method, we have used an in-house implementation whose parameter settings have been set as described in Lazebnik & Raginsky (2009). For the standard prototype approach and InfoLoss, as in our method, we have used a softmax classifier so as to allow a direct representation-quality comparison via the classification rates for our approach and those yielded by the alternatives. In Table 2, we show the accuracy of the methods evaluated along with two standard errors. As before, *probabilistic prototype* refers to our method and *standard prototype* refers to the representation computed via the direct application of k-means for learning the centers. From the table, we can conclude that our method outperforms the alternatives by delivering a mean categorization rate of 81.25%.

### 4.4. Test Likelihoods on MPEG-7 and SPECTRAL

Now, we turn our attention to the evaluation of the methods from a probabilistic point of view by reporting the likelihood of the models on the test data. In Figure 2(a), we show the test data log-likelihood for our probabilistic prototype approach and the baseline. The bar corresponds to the mean across the five trials and the segments account for two standard errors. Note that the log-likelihood for our approach is significantly greater than the one yielded by the standard approach. This is since the performance measures above account for the correctness of the labels yielded by the classifier devoid of their probability of error. Thus, our method not only delivers better performance than the alternatives, but also provides better probability estimates. In Figure 2(b) we show similar results for the SPECTRAL dataset. As with the MPEG-7 dataset, the test data log-likelihood for the SPECTRAL dataset delivered by our approach is greater than the one yielded by the alternatives and shows less variability.

### 4.5. ADULT Dataset

Finally, we applied our method to the ADULT dataset from the UCI machine learning repository (Frank & Asuncion, 2010). This dataset has been extracted from census-based data and the original task was to predict whether a person makes over $50K a year. It contains continuous and discrete input variables including *education*, *age*, *work class* (with eight different categories), etc. In order to use this dataset as a test benchmark for our method, we have grouped people's records according to the values of the discrete variables except native country and *work class* and we have selected the latter variable as our target labels. Hence, we have an 8-dimensional output variable. We represented an input instance by the set of vectors that belong to the same group (i.e. with the same values for all the discrete features excluding native country). Similarly, we have computed soft labels ($\bar{\mathcal{P}}$) by calculating for each group the proportion of records that have the same value for the corresponding label. Finally, we have sub-sampled the data to have 4146 different instances and considered 50% for training and the other 50% for testing. The regularization parameters for our model and the baseline have been set-up via cross-validation.

The (average) KL-divergence between the true (empirical) class distribution and the predictive distribution is 1.12 bits and if the soft labels were thresholded to provide a hard class assignment the accuracy of the model would be 77%. One interesting application of our model on this dataset is the automatic "discovery" of the latent population prototypes and how these relate to the class labels under consideration. Due to space limitations, we have not included these results here and postpone their analysis to future work.

*Table 2.* Classification results for the SPECTRAL dataset. The mean classification accuracy is shown along with (±) two standard errors when using 100 data-points per class for training and the rest for testing. Our probabilistic prototype method outperforms the baseline methods.

| Probabilistic Prototype | Standard Prototype | InfoLoss (Lazebnik & Raginsky, 2009) |
|---|---|---|
| **81.25 ± 0.72 %** | 75.17 ± 0.49 % | 72.16 ± 0.82 % |

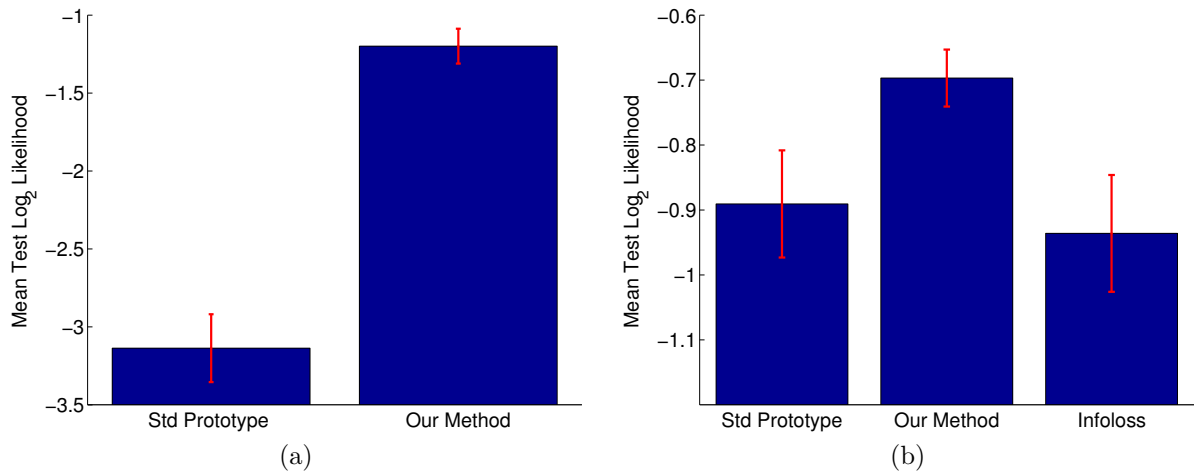

*Figure 2.* (a) The test data log likelihood on the MPEG-7 dataset, with $K = 200$ and stratified sampling for training using 50% of the data and the rest for testing. The mean values are reported along with two standard errors across 5 replications of the experiment. (b) The test data log likelihood on the SPECTRAL dataset when using 1000 data-points for training and $K = 50$.

## 5. Related Work

Generative models have been proposed as probabilistic generalizations of ad-hoc learning methods mostly for unsupervised learning scenarios (see e.g. Bishop et al., 1998). The work by Seo & Obermayer (2002) is related to ours in their attempt to generalize LVQ but their probabilistic model is inherently generative (see their equation 4). Their objective function is based on likelihood ratios and does not arise naturally from their original model.

Neural networks and, more recently, deep belief networks (DBNs) have proved popular as methods for learning latent representations with the goal of tackling difficult problems in AI (see e.g. Hinton & Salakhutdinov, 2006; Bengio & LeCun, 2007). In particular, our method can be seen as a generalization of radial basis function (RBF) networks that considers multiple probabilistic observations; applies a pooling operator over the set of vectors belonging to the same instance; and optimizes the parameters via a coordinate ascent mechanism. There is also an interesting relation between our proposed learning algorithm and how deep architectures are currently trained as we initialize our method with k-means, which can been seen as a pre-training stage, which is followed by a fine-tuning stage in DBNs.

In the computer vision community, summarization into a codebook has been used by a number of approaches in the literature. For example, Gemert et al. (2008) replace histograms with kernel density estimation (KDE) in the construction of codebooks and use the extracted features in conjunction with SVMs (non-probabilistic approach) for classification. It is interesting to mention the different between InfoLoss (Lazebnik & Raginsky, 2009) and our approach. The methods are similar in spirit but their difference is analogous to the difference between filter and wrapper methods in feature selection. While InfoLoss focuses on a filter metric (an information-theoretical objective function), our method directly includes the specific classifier in the learning process (wrapper). On a similar vein, Boureau et al. (2010) propose the learning of mid-level features for recognition in computer vision. Their work focuses on (multiple) binary classifiers and does not provide consistent probabilistic outputs across all classes. It is well-known in the machine learning theoretical literature that simple approaches to combining these classifiers such as one-against-all are inconsistent in that given optimal binary classifiers this "reduction" may not yield optimal multi-class classifiers (see e.g. Beygelzimer et al., 2009).

Finally, recent work by Bergamo et al. (2011) has proposed an approach to learning compact representations for novel category recognition. Their focus is on learning compact descriptions that can be used, for example, in image retrieval.

## 6. Conclusions

We have presented a probabilistic model for the discriminative learning of latent representations, which corresponds to a relaxed version of the popular approach of prototype-based classification. From the application viewpoint, our method can be viewed as a discriminative technique that can be used for unmixing in geosciences and remote sensing (Bergman, 2006). It can also be applied to other problems, such as population modeling where labels and proportions of these can be associated to groups of instances.

Our method requires, in general, a model for a probabilistic discriminative classifier and for purposes of illustrating the utility of our approach we have used a softmax classifier. However, our approach can, in principle, be used with any discriminative and probabilistic classifier including non-parametric methods. In the future we aim to explore the combination of non-parametric methods with our prototype learning approach and also investigate better optimization algorithms for parameter learning, e.g. based on mean field approximations.

## Acknowledgments

We thank Sarah Namin and Lars Petersson for providing us with the SPECTRAL dataset. NICTA is funded by the Australian Government as represented by the Department of Broadband, Communications and the Digital Economy and the Australian Research Council through the ICT Centre of Excellence program.

## References

Belongie, S., Malik, J., and Puzicha, J. Shape matching and object recognition using shape contexts. *IEEE TPAMI*, 24(4):509–522, 2002.

Bengio, Yoshua and LeCun, Yann. Scaling learning algorithms towards AI. In *Large-Scale Kernel Machines*. MIT Press, 2007.

Bergamo, Alessandro, Torresani, Lorenzo, and Fitzgibbon, Andrew. Picodes: Learning a compact code for novel-category recognition. In *NIPS 24*. 2011.

Bergman, M. Some unmixing problems and algorithms in spectroscopy and hyperspectral imaging. In *Applied Imagery and Pattern Recognition Workshop*, 2006.

Beygelzimer, Alina, Langford, John, and Ravikumar, Pradeep. Error-correcting tournaments. In *International conference on Algorithmic Learning Theory*, 2009.

Bishop, Christopher M., Svensén, Markus, and Williams, Christopher K. I. GTM: The generative topographic mapping. *Neural Computation*, 10(1):215–234, 1998.

Boureau, Y-Lan, Bach, Francis, LeCun, Yann, and Ponce, Jean. Learning mid-level features for recognition. In *CVPR*, 2010.

Fei-Fei, L. and Perona, P. A Bayesian hierarchical model for learning natural scene categories. In *CVPR*, 2005.

Frank, A. and Asuncion, A. UCI machine learning repository, 2010. URL **http://archive.ics.uci.edu/ml**.

Gemert, Jan C., Geusebroek, Jan-Mark, Veenman, Cor J., and Smeulders, Arnold W. Kernel codebooks for scene categorization. In *ECCV*, 2008.

Hinton, Geoffrey and Salakhutdinov, Ruslan. Reducing the dimensionality of data with neural networks. *Science*, 313(5786):504 – 507, 2006.

Kohonen, Teuvo. Improved versions of learning vector quantization. In *International Joint Conference on Neural Networks*, pp. 545–550, 1990.

Latecki, L. J., Lakamper, R., and Eckhardt, U. Shape descriptors for non-rigid shapes with a single closed contour. In *CVPR*, 2000.

Lazebnik, Svetlana and Raginsky, Maxim. Supervised learning of quantizer codebooks by information loss minimization. *IEEE TPAMI*, pp. 1294–1309, 2009.

Rosenblatt, Frank. *Principles of neurodynamics; perceptrons and the theory of brain mechanisms*. Spartan Books, 1962.

Rumelhart, D.E., Hinton, G.E., and Williams, R.J. Learning representations by back-propagating errors. *Nature*, 323(6088):533–536, 1986.

Scholkopf, Bernhard and Smola, Alexander J. *Learning with Kernels: Support Vector Machines, Regularization, Optimization, and Beyond*. MIT Press, 2001.

Seo, Sambu and Obermayer, Klaus. Soft learning vector quantization. *Neural Computation*, 15:1589–1604, 2002.

Xie, J., Heng, P., and Shah, M. Shape matching and modeling using skeletal context. *Pattern Recognition*, 41(5):1756–1767, 2008.